\documentclass[conference]{IEEEtran}
\IEEEoverridecommandlockouts

\usepackage{cite}
\usepackage{amsmath,amssymb,amsfonts}
\usepackage{graphicx}
\usepackage{textcomp}
\usepackage{xcolor}
\usepackage{algorithm}
\usepackage{algpseudocode}
\usepackage{bm}
\newcommand{\bbtheta}{\bm{\theta}}
\newcommand{\bbSig}{\bm{\Sigma}}
\newcommand{\bbphi}{\boldsymbol{\phi}}

\def\BibTeX{{\rm B\kern-.05em{\sc i\kern-.025em b}\kern-.08em
    T\kern-.1667em\lower.7ex\hbox{E}\kern-.125emX}}
\begin{document}

\title{Online scalable Gaussian processes with  conformal \\ prediction for guaranteed coverage
\thanks{J. Xu and Q. Lu are supported by NSF CAREER \#2340049; G. B. Giannakis is supported by NSF grants \# 2102312,
2103256, 1901134, 2126052, and 2128593.}
}

\author{\IEEEauthorblockN{Jinwen Xu}
\IEEEauthorblockA{\textit{School of ECE} \\
\textit{University of Georgia}\\
Athens, GA 30602, USA \\
jinwen.xu@uga.edu}
\and
\IEEEauthorblockN{Qin Lu}
\IEEEauthorblockA{\textit{School of ECE} \\
\textit{University of Georgia}\\
Athens, GA 30602, USA  \\
qin.lu@uga.edu}

\and
\IEEEauthorblockN{Georgios B. Giannakis}
\IEEEauthorblockA{\textit{Dept. of ECE} \\
\textit{University of Minnesota}\\
Minneapolis, MN 55455, USA  \\
georgios@umn.edu}

}
\maketitle

\begin{abstract}
The Gaussian process (GP) is a  Bayesian nonparametric paradigm that is widely adopted for uncertainty quantification (UQ) in a number of safety-critical applications, including robotics, healthcare, as well as surveillance. The consistency of the resulting uncertainty values however, hinges on the premise that the learning function conforms to the properties specified by the GP model, such as smoothness, periodicity and more, which may not be satisfied in practice, especially with data arriving on the fly. To combat against such model mis-specification, we propose to wed the GP with the prevailing conformal prediction (CP),  a distribution-free post-processing framework that produces {\it prediction sets} with a provably valid coverage under the sole assumption of data exchangeability. However, this assumption is usually violated in the 
 online setting, where a prediction set is sought before revealing the true label. To ensure long-term coverage guarantee, we will adpatively set the key threshold parameter based on the feedback whether the true label falls inside the prediction set. Numerical results demonstrate the merits of the online GP-CP approach relative to existing alternatives in the long-term coverage performance.
\end{abstract}

\begin{IEEEkeywords}
Uncertainty Quantification, Conformal Prediction, Online Gaussian Processes, Model Mis-specification
\end{IEEEkeywords}

\section{Introduction}
\label{sec:intro}
Uncertainty quantification (UQ) is of great importance for safety-critical applications, including robotics, healthcare, as well as surveillance. The Gaussian process (GP) is a well-established Bayesian nonparametric paradigm that provides well-calibrated uncertainty values as long as the learning function conforms to the properties specified by the GP model, such as smoothness, periodicity and more~\cite{Rasmussen2006gaussian}. In practise though, the modeling assumptions in GPs may not be satisfied, especially with data arriving on the fly. To combat against such model mis-specification, we propose to wed the online GPs with the prevailing conformal prediction (CP) framework,  a {\it distribution-free} post-processing approach that produces prediction sets with provably valid coverage for any pre-specified miscoverage rate as long as data are (pseudo-)exchangeable~\cite{vovk2005algorithmic,lei2018distribution}.

\vspace*{0.1cm}
\noindent {\bf Related Works.} 
Bayesian methods are predominant for UQ in machine learning. Thanks to its sample efficiency and closed-form expression of the posterior probability density function (pdf) of the learning function, the GP is a well-received Bayesian nonparametric framework for UQ~\cite{Rasmussen2006gaussian}. In spite of this, strong stationarity and Gaussianity assumptions in the GP limits its capability to cope with the general nonstationary functions. Alternatively, Bayesian deep learning can model arbitrary nonstationary functions owing to the representational power of deep neural networks (DNNs), but incurs high computational complexity in model training and posterior inference~\cite{wilson2020bayesian}. Also, a prior pdf has to be specified for the weights of the DNN.  Aiming to combining the merits of these two frameworks, deep kernel learning leverages a DNN to obtain a feature mapping of the original input before sending it to the GP model~\cite{wilson2016deep}. Still, it is highly susceptible to model mis-specification. 

CP, on the other hand, makes minimal statistical assumption of the learning function. As a post-processing method, CP can be coupled with any learning model and can provide valid coverage set as long as data are (pseudo)exchangeable~\cite{vovk2005algorithmic,lei2018distribution,angelopoulos2023conformal,shafer2008tutorial}. Specifically, CP relies on the so-termed {\it score} function that assesses the degree of conformality a candidate label is to its prediction, so as to determine whether the label should be assigned to the prediction set. With the score function being the negative predictive log-likelihood, conformal Bayes has also emerged that relies on the CP to adjust the prediction sets produced by Bayesian methods so as to yield the predetermined coverage probability~\cite{fong2021conformal}. Recently, CP has been adapted to the GP framework~\cite{stanton2023bayesian, papadopoulos2024guaranteed} to mitigate the strong modeling assumptions. To ensure valid coverage guarantees, the condition of data exchangeability is necessitated, which however, is easily violated with data arriving on the fly. While CP with distributional shift has been investigated~\cite{gibbs2021adaptive,gibbs2024conformal,bhatnagar2023improved,zaffran2022adaptive,barber2023conformal}, how to enable online conformalized GP with scalability and robustness to distributional shift has not been touched.

\vspace*{0.1cm}
\noindent{\bf Contributions.} Building on the aforementioned existing works, the present paper puts forth an online GP-based conformal predictor that mitigates the issue of model mis-specification in the vanilla GP. To ensure computational scalability with online continuous arrival of data, we rely on the spectral features of the GP's kernel function to obtain a linear parametric model. To further encourage valid coverage guarantees with robustness to distributional shift, the feedback that whether the true label falls into the prediction set will be leveraged to adaptively tune the key parameter in the CP. Numerical results demonstrate that the enhanced coverage performances of the resulting online GP-CP predictor relative to the plain GP-based Bayes credible set predictor and the standard CP.
\section{Preliminaries}
\subsection{Gaussian processes}
A plethora of tasks in ML boil down to learning a function $f$ that connects the $d$-dimensional input ${\bf x}$ with real-valued output $y$, visualized as ${\bf x}\rightarrow f({\bf x})\rightarrow y$. As a well-established framework to learn functions with UQ, the GP assumes a GP prior for $f$, denoted as $f\sim \mathcal{GP}(0, \kappa(\mathbf{x},\mathbf{x}'))$, where $\kappa(\cdot,\cdot)$ is a positive-definite kernel  measuring pairwise similarity between any two inputs, $\mathbf{x}$ and $\mathbf{x}'$. For any number of inputs $\mathbf{X}_t := \left[\mathbf{x}_1, \ldots, \mathbf{x}_t\right]$, the joint prior pdf of the function evaluations $\mathbf{f}_t := [f(\mathbf{x}_1),\ldots, f(\mathbf{x}_t)]^\top$ is $
p(\mathbf{f}_t|\mathbf{X}_t) = \mathcal{N} (\mathbf{f}_t ; {\bf 0}_t, {\bf K}_t )$, where ${\bf K}_t$ is a $t\times t$ covariance matrix with $(i,j)$th entry
$[{\bf K}_t]_{ij} = {\rm cov} (f(\mathbf{x}_i), f(\mathbf{x}_j)):=\kappa(\mathbf{x}_i, \mathbf{x}_j)$.

To estimate $f$, we rely on the labels $\mathbf{y}_t := [y_1, \ldots, y_t]^\top$ that are linked with $\mathbf{f}_t$ via the conditional likelihood $p (\mathbf{y}_t| \mathbf{f}_t, \mathbf{X}_t)  = \prod_{\tau = 1}^{t} p(y_{\tau}| f(\mathbf{x}_{\tau}))$. For regression, the per-datum likelihood is $p(y_{\tau}| f(\mathbf{x}_{\tau})) = {\cal N}(y_{\tau};f(\mathbf{x}_{\tau}), \sigma_n^2)$, which corresponds to the observation model $y_\tau = f({\bf x}_\tau) + n_\tau$, where $n_\tau$ is the additive white Gaussian noise with variance $\sigma_n^2$.

Given ${\cal D}_t:=\{{\bf X}_t$, ${\bf y}_t\}$, and a new test input ${\bf x}$, the goal is to find the predictive pdf for the test output $y$. Towards this, we will first write the joint pdf based on the GP prior and the Gaussian likelihood ($\mathbf{k}_{t} ({\bf x}) := [\kappa(\mathbf{x}_1, \mathbf{x}), \ldots, \kappa(\mathbf{x}_t,  \mathbf{x})]^\top$)\vspace*{-0.1cm}
\begin{align}
  \begin{bmatrix}
    {\bf y}_t\\ y  
  \end{bmatrix} \sim
  \mathcal{N}\left({\bf 0}_{t+1},
\begin{bmatrix}
  {\bf K}_t &  \mathbf{k}_{t}({\bf x})\\
  \mathbf{k}_{t}^\top({\bf x}) & \kappa (\mathbf{x}, \mathbf{x}) +\sigma_n^2
\end{bmatrix}
  \right)\;.
\end{align}
Upon conditioning on ${\cal D}_t$, one can readily obtain the predictive pdf for $y$ as \vspace*{-0.3cm}
\begin{align}
p(y|{\cal D}_t,\mathbf{x}) = \mathcal{N}(y_;\  \hat{y}_{t}({\bf x}), \sigma_{t}^2 ({\bf x})) \label{eq:y_pre}
\end{align} \vspace*{-0.5cm}\\
where the mean and variance are given as~\cite{Rasmussen2006gaussian}\vspace*{-0.2cm}
\begin{subequations}
\begin{align}	
\hat{y}_{t}({\bf x}) & = \mathbf{k}_{t}^{\top}({\bf x}) (\mathbf{K}_t + \sigma_n^2
 \mathbf{I}_t)^{-1} \mathbf{y}_t \\
\sigma_{t}^2({\bf x})& = \!\kappa(\mathbf{x},\mathbf{x})\! -\! \mathbf{k}_{t}^{\top}({\bf x}) (\mathbf{K}_t\! +\! \sigma_n^2 \mathbf{I}_t)^{-1} \mathbf{k}_{t}({\bf x}) \!+ \sigma_n^2. 
\end{align}\label{eq:plain_gpp}
\end{subequations}\vspace*{-0.5cm}\\
Having available the predictive pdf~\eqref{eq:y_pre} for $y$, one can not only obtain a {\it point} prediction $\hat{y}_{t}({\bf x})$ for $y$, but also the {\it Bayes $\beta$-credible prediction set} ${\cal K}_t^\beta ({\bf x})$ that self-assesses the quality of the prediction. When $\beta = 95\%$, the Bayes credible set based on the GP model is given by ${\cal K}_t^\beta ({\bf x}) :=\left[\hat{y}_{t}({\bf x})-2\sigma_{t}({\bf x}), \hat{y}_{t}({\bf x})+2\sigma_{t}({\bf x})\right]$. Notably,  larger prediction sets indicates highly uncertain predictions, and the vice versa. However, the consistency of the Bayes credible set ${\cal K}_t^\beta ({\bf x})$ is contingent on the model fit. If the data do not match well with the GP assumptions, the coverage of the prediction set will not be consistent~\cite{stanton2023bayesian}. To combat against such model mis-specification, we will rely on the CP framework.


\subsection{Conformal prediction}
CP, on the other hand, is a distribution-free framework~\cite{chen2018discretized} for UQ that is compatible with any prediction model~\cite{vovk2005algorithmic,lei2018distribution,angelopoulos2023conformal,angelopoulos2024conformalrisk}. 
 Suppose that we are given a prediction model $p(y|{\cal D}_t, {\bf x})$ trained on ${\cal D}_t$, which could be the aforementioned GP model~\eqref{eq:y_pre} or any other (non-)Bayesian  model~\cite{angelopoulos2021uncertainty, barber2021predictive}. To proceed, CP will rely on a negatively-oriented {\it conformity} score 
$s_t({\bf x},y): {\cal X}\times {\cal Y}\rightarrow \mathbb{R}$, which measures how well the prediction produced by the fitted model based on ${\cal D}_t$ conforms with the true value of $y$ ~\cite{Papadopoulos2008Normalized}. Specifically, a {\it larger} score indicates  significant {\it disagreement} of the prediction with the true label $y$. Upon inverting the score function, one can obtain the conformal prediction set for $y$ as
\begin{align}
    &\mathcal{C}_{t}({\mathbf{x}})= \{y\in {\cal Y}: s_{t}(\mathbf{x},y)\leq q_{t}\}\; \label{eq:CP_set}
\end{align} 
where $q_{t}$ is an estimated $1-\alpha$ quantile for the distribution of the score $s_{t} ({\bf x}, y)$. In standard CP, $q_{t}$ is set as $\lceil (1-\alpha)(t+1) \rceil$  smallest of  $\{s_t (\mathbf{x}_\tau,y_\tau)\}_{\tau=1}^t$~\cite{vovk2005algorithmic}. If $\left(\mathbf{x}_1, y_1\right), \ldots,\left(\mathbf{x}_t, y_{t}\right)$ are exchangeable, the prediction set~\eqref{eq:CP_set} enjoys the coverage guarantee:
$\mathbb{P}(y \in \mathcal{C}_t (\mathbf{x})) \geq 1-\alpha$.
 
In spite of the appealing coverage guarantee, the exchangeability assumption is often violated in practise, especially in the online setting where data samples arrive sequentially. In addition, since the prediction model is trained on ${\cal D}_t$, the conformity score evaluated at the training sample $( {\bf x}_\tau, y_\tau) (\tau \leq t)$ is usually smaller than that at any test point $\{{\bf x}, y\}$, thus invalidating the exchangeability of the scores and further leading to undercoverage~\cite{angelopoulos2023conformal}. Towards addressing these issues, we will devise a mechanism to adjust $q_t$ adaptively in the ensuing section.

\section{Online scalable GP with CP}
\label{sec:GP-CP}
Our goal here is to develop a conformalized GP predictor that can yield prediction sets with coverage guarantees in the online setting. In this context, the GP prediction model will suffer from inscalability as inverting the $t\times t$ kernel matrix in~\eqref{eq:plain_gpp}  incurs the complexity of $\mathcal{O} (t^3)$, which will become prohibitively high as $t$ grows. To effect scalability, we will first leverage a parametric GP approximant, as delineated next.

\subsection{Scalable GP with the random feature approximation}\label{sec:GP-RF}
Various attempts have been made to effect scalability in GP-based learning; see, e.g., \cite{liu2020gaussian}. Most existing approaches amount to summarizing the training data via a much smaller number ($m$) of pseudo data with inducing inputs, thereby obtaining a training-set-dependent low-rank approximant of $\mathbf{K}_t$~\cite{quinonero2005unifying}. Targeting a low-rank approximant that is not dependent on the training set, we rely here on a standardized shift-invariant $\bar{\kappa} (\cdot)$, whose inverse Fourier transform is  \vspace*{-0.2cm}
\begin{align}
\bar{\kappa}(\mathbf{x}, \mathbf{x}') &= \bar{\kappa}(\mathbf{x}-\mathbf{x}') = \int \pi_{\bar{\kappa}} (\mathbf{v}) e^{j\mathbf{v}^\top (\mathbf{x} - \mathbf{x}')} d\mathbf{v}\nonumber\\
& := \mathbb{E}_{\pi_{\bar{\kappa}}} \left[e^{j\mathbf{v}^\top (\mathbf{x} - \mathbf{x}')} \right] \label{eq:kernel_psd}
\end{align}\vspace*{-0.5cm}\\
where $\pi_{\bar{\kappa}}$ is the power spectral density (PSD), and the last equality follows after normalizing so that 
$\pi_{\bar{\kappa}}(\mathbf{v})$ integrates to $1$, what allows one to view it as a pdf.   

Upon drawing a sufficient number, say $D$, of  independent and identically distributed (i.i.d.) samples (features) $\{\mathbf{v}_i \}_{i = 1}^D$ from $\pi_{\bar{\kappa}} (\mathbf{v})$, the ensemble mean in \eqref{eq:kernel_psd} can be approximated by the sample average $\check{\bar{\kappa}}_c (\mathbf{x}, \mathbf{x}'):= \frac{1}{D} \sum_{i = 1}^D \cos (\mathbf{v}_i^\top (\mathbf{x} - \mathbf{x}'))$.

Define the real $2D\times 1$ random feature (RF) vector as~\cite{quia2010sparse}
\vspace*{-0.15cm} 
\begin{align}
&\bbphi_{\mathbf{v}} (\mathbf{x}) :=        \label{eq:phi_x}\\
&\quad \frac{1}{\sqrt{D}}\left[\sin(\mathbf{v}_1^\top \mathbf{x}), \cos(\mathbf{v}_1^\top \mathbf{x}), \ldots, \sin(\mathbf{v}_D^\top \mathbf{x}), \cos(\mathbf{v}_D^\top \mathbf{x})\right]^{\top} \nonumber
\end{align}
which allows us to replace $\check{\bar{\kappa}}_c$ with
$\check{\bar{\kappa}}(\mathbf{x}, \mathbf{x}') = \bbphi_{\mathbf{v}}^{\top} (\mathbf{x})\bbphi_{\mathbf{v}}(\mathbf{x}')
$; and thus, the parametric approximant  \vspace*{-0.2cm}
\begin{align}
{\check f} (\mathbf{x}) =  \bbphi_{\mathbf{v}}^\top (\mathbf{x}) \bbtheta, \ \  \bbtheta \sim \mathcal{N}( \mathbf{0}_{2D}, \sigma_\theta^2\mathbf{I}_{2D}) \label{eq:f_check}
\end{align}
can be viewed as coming from a realization of the Gaussian $\bbtheta$ combined with $\bbphi_{\mathbf{v}}$ to yield the GP prior with $\kappa = \sigma_\theta^2\bar{\kappa}$, where $\sigma_\theta^2$ is the magnitude of $\kappa$. 
 
With the parametric form of ${\check f} (\mathbf{x})$ in \eqref{eq:f_check}, the likelihood is also parametrized by $\bbtheta$ as $p(y_\tau|\bbtheta, \mathbf{x}_\tau) = {\cal N}(y_\tau; \bbphi_{\bf v}^{\top}({\bf x}_\tau)\bbtheta, \sigma_n^2)$. This 
together with the Gaussian prior of $\bbtheta$ (cf. \eqref{eq:f_check}), yields the Gaussian posterior ${p}(\bbtheta| \mathcal{D}_t) = \mathcal{N}(\bbtheta; \hat{\bbtheta}_{t}, \bbSig_{t})$ with mean $ \hat{\bbtheta}_{t}$ and covariance matrix $\bbSig_{t}$, based on which we can predict $y$ at each test input $\mathbf x$. In particular, this linear parametric model 
readily accommodates \emph{online} operation, 
where prediction of $y_{t+1}$ is due upon receiving $\mathbf{x}_{t+1}$ at the beginning of slot $t+1$, and the pdf of $\bbtheta$ is then updated after receiving $y_{t+1}$ at the end of slot $t+1$. 

Given $\mathbf{x}_{t+1}$, the predictive pdf for $y_{t+1}$ is obtained as\vspace*{-0.15cm}
\begin{align}
  {p}(y_{t+1}|\mathcal{D}_t, \mathbf{x}_{t+1}) = \mathcal{N}(y_{t+1}; \hat{y}_{t+1|t}, \sigma_{t+1|t}^2)  \label{eq:y_pred_RF}  
\end{align}
where the predictive mean and variance are
 $\hat{y}_{t+1|t} =  \bbphi_{\mathbf{v}}^\top(\mathbf{x}_{t+1}) \hat{\bbtheta}_{t}$ and $\sigma_{t+1|t}^2 = \bbphi_{\mathbf{v}}^\top (\mathbf{x}_{t+1})\bbSig_{t}\bbphi_{\mathbf{v}}(\mathbf{x}_{t+1})+\sigma_n^2$, respectively. Thus, one can readily obtain the following Bayes credible set for $y_{t+1}$\vspace*{-0.15cm}
 \begin{align}
   {\cal K}_t^\beta({\bf x}_{t+1}) = [\hat{y}_{t+1|t}- c_\beta\sigma_{t+1|t}, \hat{y}_{t+1|t}+ c_\beta\sigma_{t+1|t} ] \label{eq:Bayes_Credible_Set}
 \end{align}
 where $c_\beta$ is chosen such that $\int_{y\in {\cal K}_t^\beta ({\bf x}_{t+1})} \! {p}(y|\mathcal{D}_t, \mathbf{x}_{t+1}) dy \!=\! \beta$.

 Upon receiving $y_{t+1}$, the posterior pdf of $\bbtheta$ will be propagated using Bayes' rule as \vspace*{-0.1cm}
\begin{align}
{p}(\bbtheta|{\cal D}_{t+1}) &= \frac{{p}(\bbtheta| {\cal D}_{t}) {p}(y_{t+1}|\bbtheta, \mathbf{x}_{t+1})}{{p}(y_{t+1}|{\cal D}_{t},\mathbf{x}_{t+1})}  = \mathcal{N}(\bbtheta;\hat{\bbtheta}_{t+1},\bbSig_{t+1}) \label{eq:theta_up}
\end{align}\vspace*{-0.45cm}\\
whose mean and covariance are updated across slots as~\cite{lu2022incremental}~\vspace*{-0.2cm}
\begin{subequations} \label{eq:posterior_update}
\begin{align}
\hspace*{-0.2cm}
 \hat{\bbtheta}_{t+1} & = \hat{\bbtheta}_{t} + \sigma_{t+1|t}^{-2} \bbSig_{t} \bbphi_{\mathbf{v}}(\mathbf{x}_{t+1}) (y_{t+1} - \hat{y}_{t+1|t})  \nonumber \\
\hspace*{-0.5cm}
\bbSig_{t+1} & = \bbSig_{t} - {\sigma}_{t+1|t}^{-2} \bbSig_{t} \bbphi_{\mathbf{v}}(\mathbf{x}_{t+1}) \bbphi_{\mathbf{v}}^\top(\mathbf{x}_{t+1}) \bbSig_{t}\:.  \nonumber
\end{align}
\end{subequations}
Here, the per-slot complexity is only $\mathcal{O}( (2D)^2)$, significantly more scalable than the vanilla GP with cubic complexity~\eqref{eq:plain_gpp}. Further,
the resulting online scalable (OS) GP approach  is {\it memoryless} that requires no storage of past data.

\subsection{OS-GP with CP}
In the Bayes credible set~\eqref{eq:Bayes_Credible_Set}, the value of $c_\beta$ is constant over time. In the case of model mis-specification, the set ${\cal K}_\beta ({\bf x}_{t+1})$ might have poor coverage, that is, $\mathbb{P}(y_{t+1}\in {\cal K}_\beta ({\bf x}_{t+1}))\neq \beta$. Moreover, when data exchangeability is broken, the prediction set constructed by standard CP no longer has theoretical guarantees. To mitigate such inconsistency and enable provably convergent coverage, we will wed OS-GP with CP, abbreviated as ``OS-GP-CP" hereafter.

Considering the {\it Bayesian} nature of the GP predictor, the score function is chosen to be the negative predictive log-likelihood in~\eqref{eq:y_pred_RF}, namely,
$s_t({\bf x}, y) := -\log p(y|\mathcal{D}_t,\mathbf{x})$. For a new test input $\mathbf{x}_{t+1}$ at slot $t+1$, one can obtain the conformal Bayes prediction set as\vspace*{-0.25cm}
\begin{align}
 \mathcal{C}_{t}(\mathbf{x}_{t+1})&= \{y\in {\cal Y}: s_{t}(\mathbf{x}_{t+1},y)\leq q_{t}\} \nonumber\\
 &= \left[ \hat{y}_{t+1|t}- c_{t+1}\sigma_{t+1|t}, \hat{y}_{t+1|t}+ c_{t+1}\sigma_{t+1|t}\right] \label{eq:OS-GP-CP_set}
\end{align}\vspace*{-0.45cm}\\
where, unlike $c_\beta$ in~\eqref{eq:Bayes_Credible_Set}, $c_{t+1}=\sqrt{2q_{t}-\log(2\pi \sigma_{t+1|t}^2)}$ changes over time, adapting to new data samples.

Upon receiving the true label $y_{t}$, we will not only update the posterior of $\bbtheta$ in \eqref{eq:theta_up}, but also the value of $q_{t+1}$ based on the feedback that whether $y_{t+1}$ is covered by the prediction set $ \mathcal{C}_t ({\mathbf{x}_{t+1}})$. Specifically, $q_{t+1}$ is updated with the rule~\cite{gibbs2021adaptive}\vspace*{-0.2cm}
\begin{align}
q_{t+1}=q_t+\eta_t\left(\mathbb{I}(y_{t+1}\notin \mathcal{C}_{t}(\mathbf{x}_{t+1}))-\alpha\right) \label{eq:q_update}
\end{align}\vspace*{-0.5cm}\\
where $\mathbb{I}(\cdot)$ is an indicator function, which takes the value of $1$ ($0$)  if the statement inside is true (false). Intuitively, if the label $y_{t+1}$ is not covered by $\mathcal{C}_{t}(\mathbf{x}_{t+1})$, the value of $q_t$ will be increased. This update rule~\eqref{eq:q_update} can also be interpreted as the online (sub)gradient descent on the quantile loss $\rho_{1-\alpha}(u)=(1-\alpha) \max \{u, 0\}+$ $\alpha \max \{-u, 0\}$ as~\cite{gibbs2021adaptive}\vspace*{-0.25cm}
\begin{align}
    q_{t+1}=q_t-\eta_t \nabla \rho_{1-\alpha}\left(s_{t}(\mathbf{x}_{t+1},y_{t+1})-q_t\right).
\end{align}\vspace*{-0.5cm}\\
The learning rate $\eta_t$ plays a pivotal role in the long-term coverage performance. Although a judiciously chosen constant learning rate ($\eta_t = \eta,\  \forall t$) can achieve guaranteed coverage asymptotically~\cite{gibbs2021adaptive}, it will yield high variability in the instantaneous coverage even when the data samples are i.i.d.~\cite{angelopoulos2024online}. To address this issue, a decaying learning rate $\eta_t \propto t^{-a}$ for some $a\in (0,1)$ can be chosen as long as the data are stationary or has slow-varying dynamics within the time horizon ~\cite{angelopoulos2024online}. In the case when the data exhibit {\it sudden} distributional shift, the learning rate has to be reset to recover coverage more quickly. How to detect these change points is critical.  As the model fit deteriorates after the change points, the size of the prediction set will become larger to reflect the higher uncertainty. Thus, we can rely on the size of the prediction set, namely, $|\mathcal{C}_{t}(\mathbf{x}_{t+1})|=2c_{t+1}\sigma_{t+1|t}$ (cf.~\eqref{eq:OS-GP-CP_set}), to detect change points. Specifically, we calculate the average of \( |\mathcal{C}_{t}({\bf x}_{t+1})| \) over a window with $W$ slots. If the average prediction set size increases over $r$ consecutive steps, we will declare a distributional shift and then reset the learning rate.

The long-run coverage of OS-GP-CP with the time-varying learning rate $\eta_t$ can be guaranteed via the following theorem~\cite{angelopoulos2024online}.

\vspace*{0.1cm}
\noindent\textbf{Theorem}. {\it For an arbitrary sequence of data points $\left(\mathbf{x}_1, y_1\right),\left(\mathbf{x}_2, y_2\right), \ldots$, and an arbitrary positive sequence of learning rates $\{\eta_t\}$, if the score function $s_t ({\bf x},y): {\cal X}\times {\cal Y}\rightarrow [0, B]$ and the initial threshold $q_0 \in[0, B]$, the long-run coverage rate of OS-GP-CP is given by
\begin{align}
    \left|\frac{1}{T} \sum_{t=0}^{T-1} \mathbb{I}({y_{t+1} \in \mathcal{C}_t\left({\bf x}_{t+1}\right)})-(1-\alpha)\right| \le  \frac{C N_T}{T}
\label{eq:gap}
\end{align}
where $C=2(B+\max _{0 \leq t \leq T-1} \eta_t)/(\min _{0 \leq t \leq T-1} \eta_t)$
and $N_T=\sum_{\tau=1}^T \mathbb{I}({\eta_\tau>\eta_{\tau-1}})$ is the number of times the learning rate is increased.  }

Thus, as long as the step size does not decay
too quickly, and the number of resets $N_T$ is sublinear in $T$, OS-GP-CP can achieve long-run coverage convergence.



\begin{figure}[ht]
    \centering \includegraphics[width=0.5\textwidth]{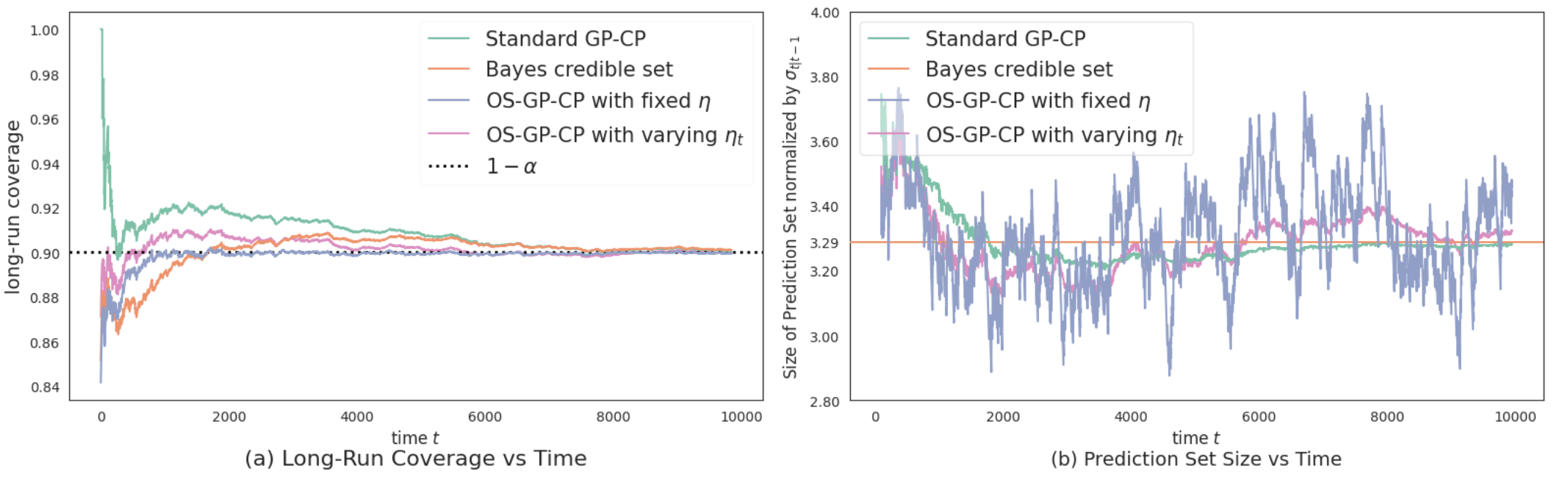}
    \caption{Synthetic dataset without distributional shift.}
    \label{fig:syn_iid}
\end{figure}
\begin{figure}[ht]
    \centering
\includegraphics[width=0.5\textwidth]{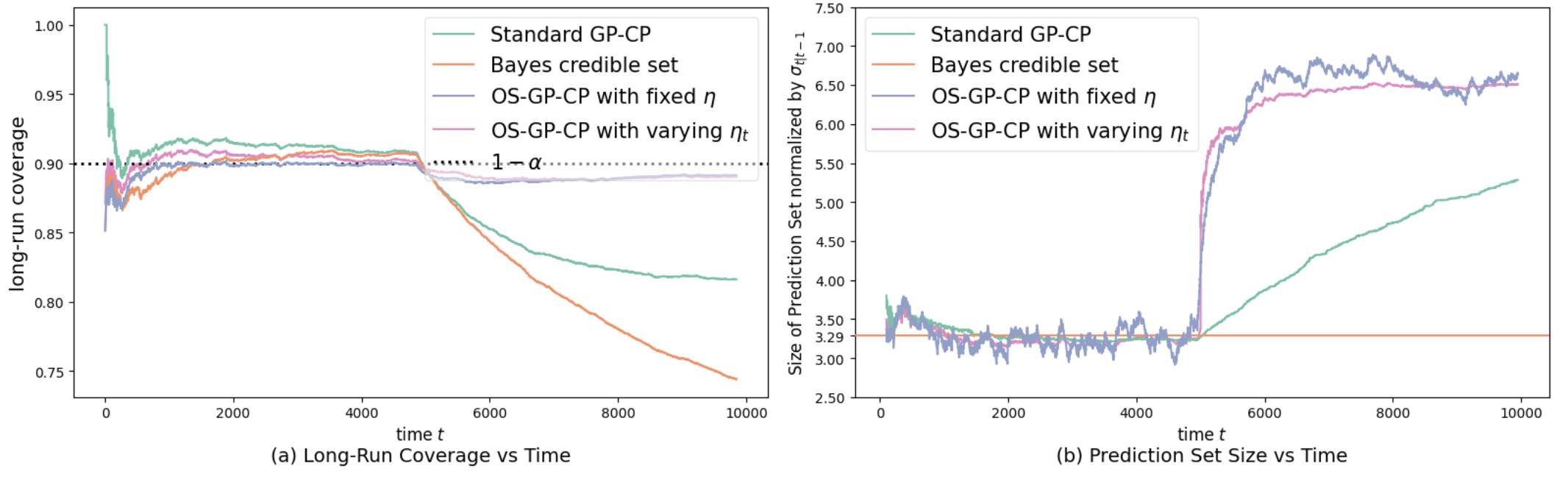}
    \caption{Synthetic dataset with distributional shift.}
    \label{fig:syn_non_iid}
\end{figure}
\begin{figure}[ht]
    \centering
\includegraphics[width=0.5\textwidth]{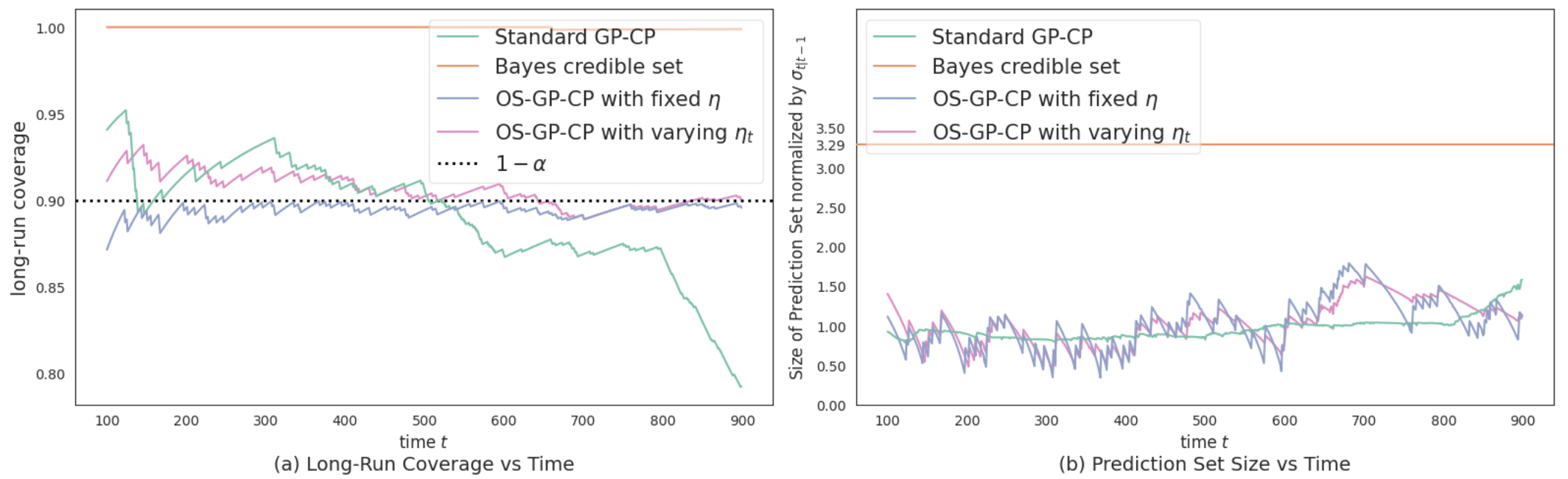}
    \caption{Real stock price dataset.}
    \label{fig:real_apple}
\end{figure}

\section{NUMERICAL EXPERIMENTS}
\label{sec:numerical_results}
This section assesses the coverage performance of the proposed OS-GP-CP approach with a constant $\eta = 0.05$ and time-varying $\eta_t$ through numerical experiments. For the latter, we leverage on the approach in Sec.~III-B with window size $W=15$ and $r=100$ to detect the change points, at which the learning rate is reset. During any two change points, a decaying learning rate $\eta_t  = t^{-3/5}$ is adopted.


For comparison, the competing alternatives consist of the Bayes credible set~\eqref{eq:Bayes_Credible_Set} with $\beta = 1-\alpha$ and standard CP set~\eqref{eq:CP_set} with $q_t$ estimated as $1-\alpha$ quantile of past scores.
For the GP model, the kernel is given by the radial basis function (RBF) $\kappa(\mathbf{x},\mathbf{x}') = \sigma_\theta^2 \exp (-\|\mathbf{x}-\mathbf{x}'\|_2^2/\sigma_l^2)$, where the kernel magnitude $\sigma_\theta^2$ and characteristic lengthscale $\sigma_l^2$ are optimized together with the noise variance $\sigma_n^2$ by maximizing the marginal likelihood using the first $100$ data points. In the RF approximation, the number $D$ of spectral features is $200$. 
The competing methods are tested on three datasets, namely, a synthetic one with i.i.d. (exchangeable) samples, a synthetic one with distributional shift, and real-world ~\cite{yfinance}. The performance metrics are the long-run average coverage up to slot $t$, given by $1/t \sum_{\tau=0}^{t-1} \mathbb{I}(y_{\tau+1} \in {\cal C}_\tau (\mathbf{x}_{\tau+1}))$, and the size of the prediction set. The value of $\alpha$ is set to be $0.1$ in all the tests.

\vspace*{0.1cm}
\noindent {\bf Synthetic data with i.i.d. samples.} $10,000$ samples are generated based on $ y_t = \sin(x_t) + n_t$, where the input feature $x_t \sim {\cal U}(0,10) $ and $n_t \sim {\cal N}(0, 0.1^2)$. In this idea setting with i.i.d. (exchangeable) samples, the coverage of all the prediction sets converges to $1-\alpha = 0.9$ as observed in Fig.~\ref{fig:syn_iid}. While the size of standard CP set converges to that of the Bayes credible set, the size of OS-GP-CP with constant $\eta$  has noticeable oscillation, corroborating the high variation in the instantaneous coverage~\cite{angelopoulos2024online}. On the other hand, OS-GP-CP with varying $\eta_t$ produces prediction sets with more stable size.


\vspace*{0.1cm}
\noindent {\bf Synthetic data with distributional shift.} The second dataset, also consisting of $10,000$ data points, was generated using $y_t = \sin(\mathbf{x}_t) + n_t \mathbb{I}({t \leq 5000}) + \epsilon_t \mathbb{I}({t > 5000})$, where $n_t \sim {\cal N}(0, 0.1^2)$ and $\epsilon_t \sim {\cal N}(0, 0.2^2)$. Apparently, there is a distributional shift at slot $5000$. As shown in Fig.~\ref{fig:syn_non_iid}, the Bayes credible set shows a marked coverage drop after 5000 points due to model mis-specification. Similarly, the coverage for standard CP also falls below 0.9 as it struggles with non-exchangeable data. OS-GP-CP with both fixed and varying step sizes quickly recover coverage around $0.9$. Note that in the latter, a change point is detected and the learning rate is reset at slot $5003$, validating the efficacy of the change point detection mechanism in Sec.III-B.

\vspace*{0.1cm}
\noindent {\bf Real stock price time-series data.} This dataset tracks the closing stock price ($y_t$) of Apple Inc. (AAPL) using its opening, high, and low prices, collected in ${\bf x}_t$, from January 4, 2016, to July 31, 2019~\cite{yfinance}. Here, continuous distribution changes over time can be observed in the data samples. While the traditional Bayes credible set achieves full coverage as shown in Fig.~\ref{fig:real_apple}, it did so by excessively enlarging the prediction set. Standard CP also struggles with the evolving data distribution, leading to a drop in coverage below $0.8$. By contrast, our proposed OS-GP-CP with both constant and time-varying learning rates yield a more modest prediction set size with average coverage at $0.9$.

\section{Conclusions}\label{sec:conclusions}
This work relied on the CP framework to allow for prediction sets with valid coverage guarantees to be constructed by the GP predictor. To effect scalability with the online arrival of data, the RF-based parametric OS-GP model was adopted that has constant model update per slot. To further combat against distributional shift, we leveraged the feedback whether the true label resides in the prediction set to adaptively adjust the key threshold parameter when constructing the prediction set. Numerical results showcased that the resulting OS-GP-CP approach has improved coverage performance relative to the competing baselines under the distributional shift.

\clearpage

\bibliographystyle{IEEEtran}  
\bibliography{CP_GP}   

\begin{thebibliography}{10}
\providecommand{\url}[1]{#1}
\csname url@samestyle\endcsname
\providecommand{\newblock}{\relax}
\providecommand{\bibinfo}[2]{#2}
\providecommand{\BIBentrySTDinterwordspacing}{\spaceskip=0pt\relax}
\providecommand{\BIBentryALTinterwordstretchfactor}{4}
\providecommand{\BIBentryALTinterwordspacing}{\spaceskip=\fontdimen2\font plus
\BIBentryALTinterwordstretchfactor\fontdimen3\font minus \fontdimen4\font\relax}
\providecommand{\BIBforeignlanguage}[2]{{%
\expandafter\ifx\csname l@#1\endcsname\relax
\typeout{** WARNING: IEEEtran.bst: No hyphenation pattern has been}%
\typeout{** loaded for the language `#1'. Using the pattern for}%
\typeout{** the default language instead.}%
\else
\language=\csname l@#1\endcsname
\fi
#2}}
\providecommand{\BIBdecl}{\relax}
\BIBdecl

\bibitem{Rasmussen2006gaussian}
C.~E. Rasmussen and C.~K. Williams, \emph{Gaussian processes for machine learning}.\hskip 1em plus 0.5em minus 0.4em\relax MIT press Cambridge, MA, 2006.

\bibitem{vovk2005algorithmic}
V.~Vovk, A.~Gammerman, and G.~Shafer, \emph{Algorithmic learning in a random world}.\hskip 1em plus 0.5em minus 0.4em\relax Springer, 2005, vol.~29.

\bibitem{lei2018distribution}
J.~Lei, M.~G’Sell, A.~Rinaldo, R.~J. Tibshirani, and L.~Wasserman, ``Distribution-free predictive inference for regression,'' \emph{Journal of the American Statistical Association}, vol. 113, no. 523, pp. 1094--1111, 2018.

\bibitem{wilson2020bayesian}
A.~G. Wilson and P.~Izmailov, ``Bayesian deep learning and a probabilistic perspective of generalization,'' \emph{Advances in neural information processing systems}, vol.~33, pp. 4697--4708, 2020.

\bibitem{wilson2016deep}
A.~G. Wilson, Z.~Hu, R.~Salakhutdinov, and E.~P. Xing, ``Deep kernel learning,'' \emph{Proc. Int. Conf. Artif. Intel. and Stats.}, pp. 370--378, 2016.

\bibitem{angelopoulos2023conformal}
A.~N. Angelopoulos and S.~Bates, ``Conformal prediction: A gentle introduction,'' \emph{Foundations and Trends{\textregistered} in Machine Learning}, vol.~16, no.~4, pp. 494--591, 2023.

\bibitem{shafer2008tutorial}
G.~Shafer and V.~Vovk, ``A tutorial on conformal prediction,'' \emph{Journal of Machine Learning Research}, vol.~9, pp. 371--421, jun 2008.

\bibitem{fong2021conformal}
E.~Fong and C.~C. Holmes, ``Conformal {B}ayesian computation,'' \emph{Proc. Adv. Neural Inf. Process. Syst.}, vol.~34, pp. 18\,268--18\,279, 2021.

\bibitem{stanton2023bayesian}
S.~Stanton, W.~Maddox, and A.~G. Wilson, ``Bayesian optimization with conformal prediction sets,'' \emph{Proc. Int. Conf. Artif. Intel. and Stats.}, pp. 959--986, 2023.

\bibitem{papadopoulos2024guaranteed}
H.~Papadopoulos, ``Guaranteed coverage prediction intervals with {G}aussian process regression,'' \emph{IEEE Trans. Pattern Anal. Mach. Intel.}, 2024.

\bibitem{gibbs2021adaptive}
I.~Gibbs and E.~Candes, ``Adaptive conformal inference under distribution shift,'' \emph{Advances in Neural Information Processing Systems}, 2021.

\bibitem{gibbs2024conformal}
I.~Gibbs and E.~Candès, ``Conformal inference for online prediction with arbitrary distribution shifts,'' \emph{Journal of Machine Learning Research}, vol.~25, pp. 1--36, 2024, submitted 10/22; Revised 5/24; Published 5/24.

\bibitem{bhatnagar2023improved}
A.~Bhatnagar, H.~Wang, C.~Xiong, and Y.~Bai, ``Improved online conformal prediction via strongly adaptive online learning,'' \emph{International Conference on Machine Learning}, vol. 2023, pp. 2337--2363, 2023.

\bibitem{zaffran2022adaptive}
M.~Zaffran, O.~Féron, Y.~Goude, J.~Josse, and A.~Dieuleveut, ``Adaptive conformal predictions for time series,'' \emph{International Conference on Machine Learning}, vol. 2022, pp. 25\,834--25\,866, 2022.

\bibitem{barber2023conformal}
R.~F. Barber, E.~J. Candes, A.~Ramdas, and R.~J. Tibshirani, ``Conformal prediction beyond exchangeability,'' \emph{The Annals of Statistics}, vol.~51, no.~2, pp. 816--845, 2023.

\bibitem{chen2018discretized}
W.~Chen, K.-J. Chun, and R.~F. Barber, ``Discretized conformal prediction for efficient distribution-free inference,'' \emph{Stat}, vol.~7, no.~1, p. e173, 2018.

\bibitem{angelopoulos2024conformalrisk}
A.~N. Angelopoulos, S.~Bates, A.~Fisch, L.~Lei, and T.~Schuster, ``Conformal risk control,'' \emph{The Twelfth International Conference on Learning Representations}, 2024.

\bibitem{angelopoulos2021uncertainty}
A.~N. Angelopoulos, S.~Bates, M.~Jordan, and J.~Malik, ``Uncertainty sets for image classifiers using conformal prediction,'' \emph{International Conference on Learning Representations}, 2021.

\bibitem{barber2021predictive}
R.~F. Barber, E.~J. Candès, A.~Ramdas, and R.~J. Tibshirani, ``Predictive inference with the jackknife+,'' \emph{The Annals of Statistics}, vol.~49, no.~1, 2021.

\bibitem{Papadopoulos2008Normalized}
H.~Papadopoulos, A.~Gammerman, and V.~Vovk, ``Normalized nonconformity measures for regression conformal prediction,'' \emph{Proceedings of the 26th IASTED International Conference on Artificial Intelligence and Applications}, pp. 64--69, 2008.

\bibitem{liu2020gaussian}
H.~Liu, Y.-S. Ong, X.~Shen, and J.~Cai, ``When {G}aussian process meets big data: {A} review of scalable {GP}s,'' \emph{IEEE Trans. Neural Net. and Learn. Syst.}, vol.~31, no.~11, pp. 4405--4423, 2020.

\bibitem{quinonero2005unifying}
J.~Qui{\~n}onero-Candela and C.~E. Rasmussen, ``A unifying view of sparse approximate {Gaussian} process regression,'' \emph{Journal of Machine Learning Research}, vol.~6, pp. 1939--1959, 2005.

\bibitem{quia2010sparse}
M.~L\'azaro-Gredilla, J.~Qui\~nonero Candela, C.~E. Rasmussen, and A.~Figueiras-Vidal, ``Sparse spectrum {G}aussian process regression,'' \emph{J. Mach. Learn. Res.}, vol.~11, no. Jun, pp. 1865--1881, 2010.

\bibitem{lu2022incremental}
Q.~Lu, G.~V. Karanikolas, and G.~B. Giannakis, ``Incremental ensemble {G}aussian processes,'' \emph{IEEE Trans. Pattern Anal. Mach. Intel.}, vol.~45, no.~2, pp. 1876--1893, 2022.

\bibitem{angelopoulos2024online}
A.~N. Angelopoulos, R.~Barber, and S.~Bates, ``Online conformal prediction with decaying step sizes,'' \emph{Forty-first International Conference on Machine Learning}, 2024.

\bibitem{yfinance}
\BIBentryALTinterwordspacing
{Ran Aroussi}, ``{yfinance: Yahoo! Finance market data downloader},'' 2024, python library used for data retrieval. [Online]. Available: \url{https://github.com/ranaroussi/yfinance}
\BIBentrySTDinterwordspacing

\end{thebibliography}
\end{document}